%% file: main.tex
\documentclass[a4paper,fleqn]{cas-dc}

\usepackage[numbers]{natbib}

\def\tsc#1{\csdef{#1}{\textsc{\lowercase{#1}}\xspace}}
\tsc{WGM}
\tsc{QE}
\tsc{EP}
\tsc{PMS}
\tsc{BEC}
\tsc{DE}
\begin{document}
\let\WriteBookmarks\relax
\def\floatpagepagefraction{1}
\def\textpagefraction{.001}
\shorttitle{RNNs for
Source Identication and Forecasting of Dynamical Systems}
\shortauthors{P Saha et~al.}

\title [mode = title]{Physics-Incorporated Convolutional Recurrent Neural Networks 
for Source Identification and Forecasting of  Dynamical Systems }                      
\tnotetext[1]{This material is based on work sponsored by the Army Research Office
and was accomplished under Grant Number W911NF-19-1-0447. The views and conclusions contained in this document are those of the authors and should not be interpreted as representing the official policies, either expressed or implied, of the Army Research Office or the U.S. Government.}

\author{Priyabrata Saha}[orcid=0000-0002-6933-0660]
\cormark[1]
\ead{priyabratasaha@gatech.edu}

\author{Saurabh Dash}

\author {Saibal Mukhopadhyay}

\address{School of Electrical and Computer  Engineering, Georgia Institute of Technology, Atlanta, GA 30332, USA}

\cortext[cor1]{Corresponding author}

\begin{abstract}
Spatio-temporal dynamics of physical processes are generally modeled using partial differential equations (PDEs).  Though the core dynamics follows some principles of physics, real-world physical processes are often driven by unknown external sources.  In such cases, developing a purely analytical model becomes very difficult and data-driven modeling can be of assistance. In this paper, we present a hybrid framework combining physics-based numerical models with deep learning for source identification and forecasting of spatio-temporal dynamical systems with unobservable time-varying external sources. We formulate our model PhICNet as a convolutional recurrent neural network (RNN) which is end-to-end trainable for spatio-temporal evolution prediction of dynamical systems and learns the source behavior as an internal state of the RNN. Experimental results show that the proposed model can forecast the dynamics for a relatively long time and identify the sources as well.
\end{abstract}


\maketitle

\input{introduction}

\input{related}

\input{problem}

\input{model}

\input{experiments}

\input{conclusion}

\bibliographystyle{cas-model2-names}

\bibliography{ref}


\end{document}

%% file: introduction.tex
\section{Introduction}
\label{sec:introduction}
Understanding the behavior of dynamical systems is a fundamental problem in science and engineering. Classical approaches of modeling dynamical systems involve formulating ordinary or partial differential equations (ODEs or PDEs) based on various physical principles, profound reasoning, intuition, knowledge and verifying those with experiments and observations. However, in many modern fields, such as climate science, neuroscience, epidemiology, finance etc., analytical modeling is very challenging
and/or can only describe the system behavior partially.   
Recent successes of machine learning methods in complex sequence prediction tasks \cite{sutskever2014sequence, chung2014empirical, xingjian2015convolutional, finn2016unsupervised, kumar2016ask, lee2017desire, minderer2019unsupervised} along with development in sensor technologies and computing systems motivate to predict the evolution of dynamical system directly from observation data.
Consequently, a number of machine learning models which incorporate knowledge from physics or applied mathematics, have been introduced for modeling complex dynamical systems \cite{schaeffer2017learning,rudy2017data,raissi2018deep,long2018pde,long2018hybridnet, de2019deep}.

\begin{figure*}
  \centering
  \includegraphics[width=0.7\linewidth]{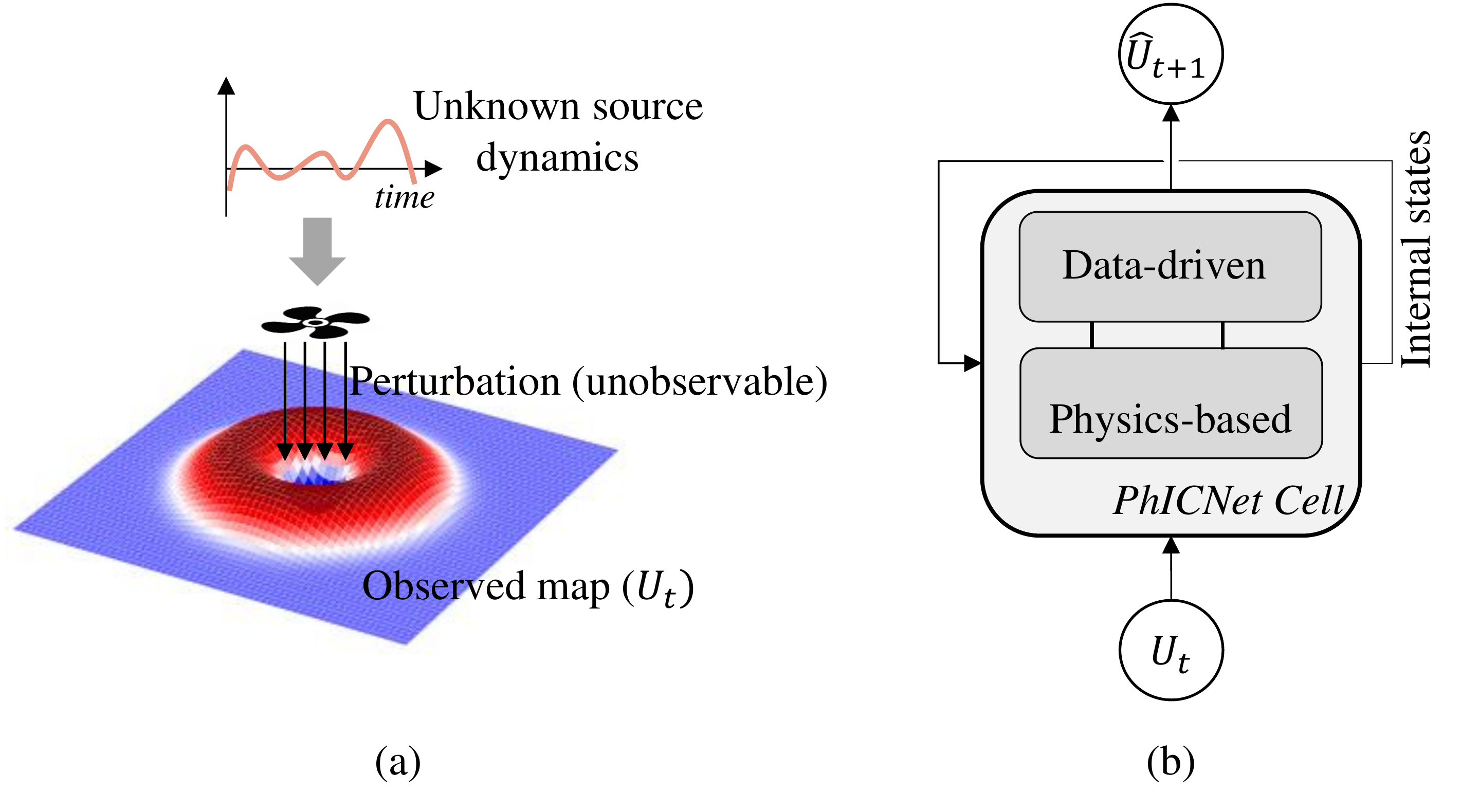}
  \caption{(a): An example dynamical system governed by partially-known PDE and unknown source dynamics. (b): High-level diagram of the recurrent network cell that is used to model such dynamics. Unobservable perturbation is learned as an internal state of PhICNet cell.}
  \label{fig:intro}
\end{figure*}

Real-world dynamical systems are often subjected to perturbation from time-varying external sources. For example, in many fields of applied sciences, such as ocean acoustics, geophysics etc., the dynamical systems are often stimulated with external sources to study the corresponding environment. Perturbation from an external source, generally referred as source term, acts as an external force or stimulation that changes the behavior of the system from its default course.
Figure \ref{fig:intro}(a) shows an example where an elastic membrane under tension is being perturbed with an independent time-varying pressure. The perturbation pressure will cause a vertical displacement on the membrane where it is applied and the membrane will respond to restore its shape due to its elasticity. Consequently, new distortions/undulations will be created in the neighboring region forming a wave that propagates across the membrane. Consider the case where the undulation of the membrane can be observed as a regularly sampled spatiotemporal sequence but the spatiotemporal variation in the source term is not observable.
Moreover, though the basic governing dynamics (wave equation) of the system is known, the physical parameters such as propagation speed in that particular medium may not be known. In such scenarios, we need to be able to predict spatiotemporal evolution of dynamical systems from partial knowledge of the governing dynamics and limited observability of the factors that influence the system behaviors. 

In this paper, we consider to model a generic PDE-based dynamical system which is perturbed with external sources that follow another independent dynamics. Our goal is to design a neural network model that can be used for long-term prediction of the entire systems, as well as of the source dynamics. We assume the physical quantity that follows the combined dynamics can be observed as regularly sampled spatiotemporal sequence, but \textit{the source or perturbation is not observable (neither during training nor testing) separately}. It is further assumed that we know what type of system we are observing and therefore scientific knowledge of such system can be incorporated in the model. In particular, we assume that the analytical form the underlying PDE is known a priori, \textit{but the physical parameters (e.g., propagation speed in the elastic membrane example) of the system are unknown}. Therefore, it is not possible to separate out the source dynamics from the PDE. Hence, the main challenge lies in the fact that the unknown source dynamics cannot be learned using direct supervision because we do not have access to the ground truth. Rather, we need to formulate a supervised learning task for the (observable) combined dynamics that will inherently learn the hidden source dynamics.    
 
We propose \textbf{Ph}ysics-\textbf{I}ncorporated \textbf{C}onvolutional Recurrent Neural \textbf{Net}works (PhICNet) that combine physical models with data-driven models to learn the behavior of dynamical systems with unobservable time-varying external source term. Figure \ref{fig:intro}(b) shows the high-level diagram of a PhICNet cell. The basic concept of PhICnet is based on two key contributions. First, a generic homogeneous PDE of any temporal order can be mapped into a recurrent neural network (RNN) structure with trainable physical parameters. Second, the basic RNN structure for homogeneous PDE can be modified and integrated with a residual encoder-decoder network to identify and learn the source dynamics. The RNN structure stores the homogeneous solution of the underlying PDE as an internal state which is then compared with the input (observation) at the next step to find out if there exists some source term or perturbation. The residual encoder-decoder network learns to propagate the source term as it is in case of constant perturbation or predict its progress in case of dynamic perturbation. The PhICNet cell stores the estimated perturbation or source term as an internal cell which can be used to understand behaviour of source dynamics. The integrated model can be trained end-to-end using stochastic gradient descent (SGD) based backpropagation through time (BPTT) algorithm. 

Few existing models in literature can be used or extended to perform the spatiotemporal sequence prediction of dynamical systems with time-varying independent source. Pure data-driven models like ConvLSTM \cite{xingjian2015convolutional}, residual networks \cite{he2016deep, mao2016image} can be used directly, but these models lack consideration of underlying physical dynamics resulting in limited accuracy. Furthermore, these pure data-driven models cannot identify the source dynamics separately. Deep hidden physics models (DHPM) \cite{raissi2018deep} can model the underlying homogeneous PDE, but does not consider any nonlinear source term. A polynomial approximation can be added in DHPM to model nonlinear internal source. This strategy is used in PDE-Net \cite{long2018pde} to incorporate nonlinear source term; however they consider only internal source term that is a nonlinear function of the observed physical quantity. DHPM, PDE-Net only consider PDEs that are first-order in time, although can be extended to model higher temporal order systems if we know the temporal order a priori. 

Our approach fundamentally differs from the past approaches as we model the system as an RNN that couples the known and unknown dynamics within the hidden cells and enables end-to-end training. We evaluate our model along with other baselines for three types of dynamical systems: a heat diffusion system, a wave propagation system and, a Burgers' fluid flow system. Experiments show that our model provides more accurate prediction compared to the baseline models. Furthermore, we show that our model can predict the source dynamics separately which is not possible with other models.

%% file: related.tex
\section{Related Work}
\label{sec:related}

\subsection{RNNs for Dynamical Systems}
Several studies have interpreted RNNs as approximation of dynamical systems \cite{funahashi1993approximation, li2005approximation, liao2016bridging, chen2018neural}. Recently, a number of RNN architectures have been proposed for data-driven modeling of dynamical systems. Trischler and D’Eleuterio \cite{trischler2016synthesis} proposed an algorithm for efficiently training RNNs to replicate dynamical systems and demonstrated its capability to approximate attractor dynamical systems. A class of RNNs, namely Tensor-RNNs, has been proposed in \cite{yu2017long, yu2017learning} for long-term prediction of nonlinear dynamical systems. Yeo and Melnyk \cite{yeo2019deep} use LSTM for 
long-term prediction of nonlinear dynamics.

\subsection{Learning PDEs from Data}
Recently, numerous attempts have been made on data-driven discovery of PDE-based dynamical systems. Schaeffer
\cite{schaeffer2017learning}, Rudy et al.\cite{rudy2017data} use sparse optimization techniques to choose the best candidates from a library of possible partial derivatives of the unknown governing equations. Raissi and Karniadakis \cite{raissi2018hidden} proposed a method to learn scalar parameters of PDEs using Gaussian process. A deep neural network is introduced in \cite{raissi2017physics} to approximate the solution of a nonlinear PDE. The predicted solution is then fed to a physics-informed neural network to validate that solution. The physics-informed neural network is designed based on the explicit form of the underlying PDE which is assumed to be known except for a few scalar learnable parameters. Raissi \cite{raissi2018deep} extended \cite{raissi2017physics} to replace the known PDE-based neural network to a generalized neural network which discovers the dynamics of underlying PDE using predicted solution and its derivatives. The inputs of the neural network are the partial derivatives up to a maximum order which is considered as a hyperparameter. Long et al. 
\cite{long2018pde} introduced PDE-Net that uses trainable convolutional filters to perform differentiations. Filters are initialized as differentiating kernels of corresponding orders, and trained by imposing some constraints to maintain differentiating property. They assumed that the maximum order of derivative is known a priori. In PDE-Net 2.0 \cite{long2019pde}, a symbolic neural network is integrated with original PDE-Net to uncover more complex analytical form. de Bezenac et al. \cite{de2019deep} proposed a convolutional neural network (CNN) that incorporates prior scientific knowledge for the problem of forecasting sea surface temperature. They design their model based on the general solution of the advection-diffusion equation. Long et al. \cite{long2018hybridnet} studied a problem similar to ours where the source or perturbation term of the PDE follows another dynamics. They mapped the known PDE into a cellular neural network with trainable physical parameters and integrate that with ConvLSTM \cite{xingjian2015convolutional} that models the source dynamics. However, they assumed that the source or perturbation is observable and they train the two networks separately.

%% file: problem.tex
\section{Problem Description}
\label{sec:problem}

We consider dynamical systems governed by the following generic inhomogeneous PDE 
\begin{align}
    \frac{\partial^n u}{\partial t^n}  &=  F \bigg( x, y, u, \frac{\partial u}{\partial x}, \frac{\partial u}{\partial y}, \frac{\partial^2 u}{\partial x^2}, \frac{\partial^2 u}{\partial x \partial y}, \frac{\partial^2 u}{\partial y^2}, \cdots; \theta \bigg) \nonumber \\
    & \ \   + v(x,y,t), \qquad (x, y) \in \Omega \subset \mathbb{R}^2, t \in [0,T].
    \label{eqn:main_dyn}
\end{align}
$u(x, y, t) \in \mathbb{R}$ is the observed physical quantity at the spatial location $(x,y) \in \Omega$ at time $t \in [0, T]$. $F$ is a linear or nonlinear function of the observed physical quantity and its spatial derivatives of different orders. $\theta$ corresponds to the physical parameters of the system. For example, in the case of wave propagation in elastic membrane (Figure \ref{fig:intro}(a)), $\theta$ corresponds to the propagation speed in that medium and, $F$ takes the following form \\ 
$F \bigg( x, y, u, \frac{\partial u}{\partial x}, \frac{\partial u}{\partial y}, \frac{\partial^2 u}{\partial x^2}, \frac{\partial^2 u}{\partial x \partial y}, \frac{\partial^2 u}{\partial x^2}, \cdots; \theta \bigg) = \bigg(\frac{\partial^2 u}{\partial x^2} + \frac{\partial^2 u}{\partial y^2}\bigg) \theta$. $v(x, y, t) \in \mathbb{R}$ is the unobservable source term or perturbation at location $(x,y) \in \Omega$ at time $t \in [0, T]$ which is governed by another independent dynamics delineated by:
\begin{equation}
    \frac{\partial^k v}{\partial t^k} = G(x, y, v)
    \label{eqn:source_dyn}
\end{equation}
Our goal is to learn the spatiotemporal evolution of the source or perturbation (i.e. $v$) as well as of the entire system jointly defined by equation \ref{eqn:main_dyn} and equation \ref{eqn:source_dyn} while observing only $u$. 

\textit{\textbf{Assumptions}} We make following assumptions about the problem:
\vspace{-2mm}
\begin{itemize}
    \item We have the \textit{a priori} knowledge about what type of physical quantities we are observing and how such system behave in absence of any external perturbation or source. In other words, we know the analytical form of function $F$ and the temporal order $n$.
    \vspace{-2mm}
    \item Physical parameters of the system $\theta$ are not known.
    \vspace{-2mm}
    \item The perturbation $v$ is \textbf{not} observable or cannot be computed directly from the observed quantity $u$ as $\theta$ is unknown. Therefore, we do not have access to any ground truth for direct supervision to learn the source dynamics. The temporal order of the perturbation or source dynamics is either known or can be chosen as a hyperparameter.   
\end{itemize}
This problem can be formulated as a spatiotempotral sequence prediction problem. Suppose the observation space is discretized into a $X \times Y$ grid and $U_t \in \mathbb{R}^{X \times Y}$ is the observed map at timestep $t \in \{0, 1, \ldots, T\}$. We aim to design a physics-incorporated convolutional-RNN $\mathcal{R}$:
\begin{align}
    \widehat{U}_{t+1} &= \mathcal{R}(\widehat{U}_t, \ldots, \widehat{U}_{t-n-k+1}), \nonumber\\ \quad t &\in \{n+k-1, \ldots, T-1\}
\end{align}
such that $\sum_{t=n+k}^{T} \mathcal{L}(U_{t}, \widehat{U}_{t})$ is minimized. $\widehat{U}_t$ is the predicted map at timestep $(t-1)$
%and $\widehat{U}_t = U_t \quad \forall \  t \in \{0, \ldots, n+k-1\} $. 
and $\mathcal{L}$ is the loss function between observed map and predicted map. For the very first prediction $\widehat{U}_{n+k}$, we use the initial $(n+k)$ true observed maps  $U_t, \  \  t \in \{0, \ldots, n+k-1\}$ as the input to $\mathcal{R}$. Note that we need $(n+k)$ previous maps, instead of just $n$, to predict the next map in an $n^{th}$ order (temporal) system because the source/perturbation is unknown and source follows a $k^{th}$ order dynamics.

%% file: model.tex
\section{The PhICNet Model: Foundation and Design}
\label{sec:model}

\begin{figure*}
  \centering
  \includegraphics[width=1\linewidth]{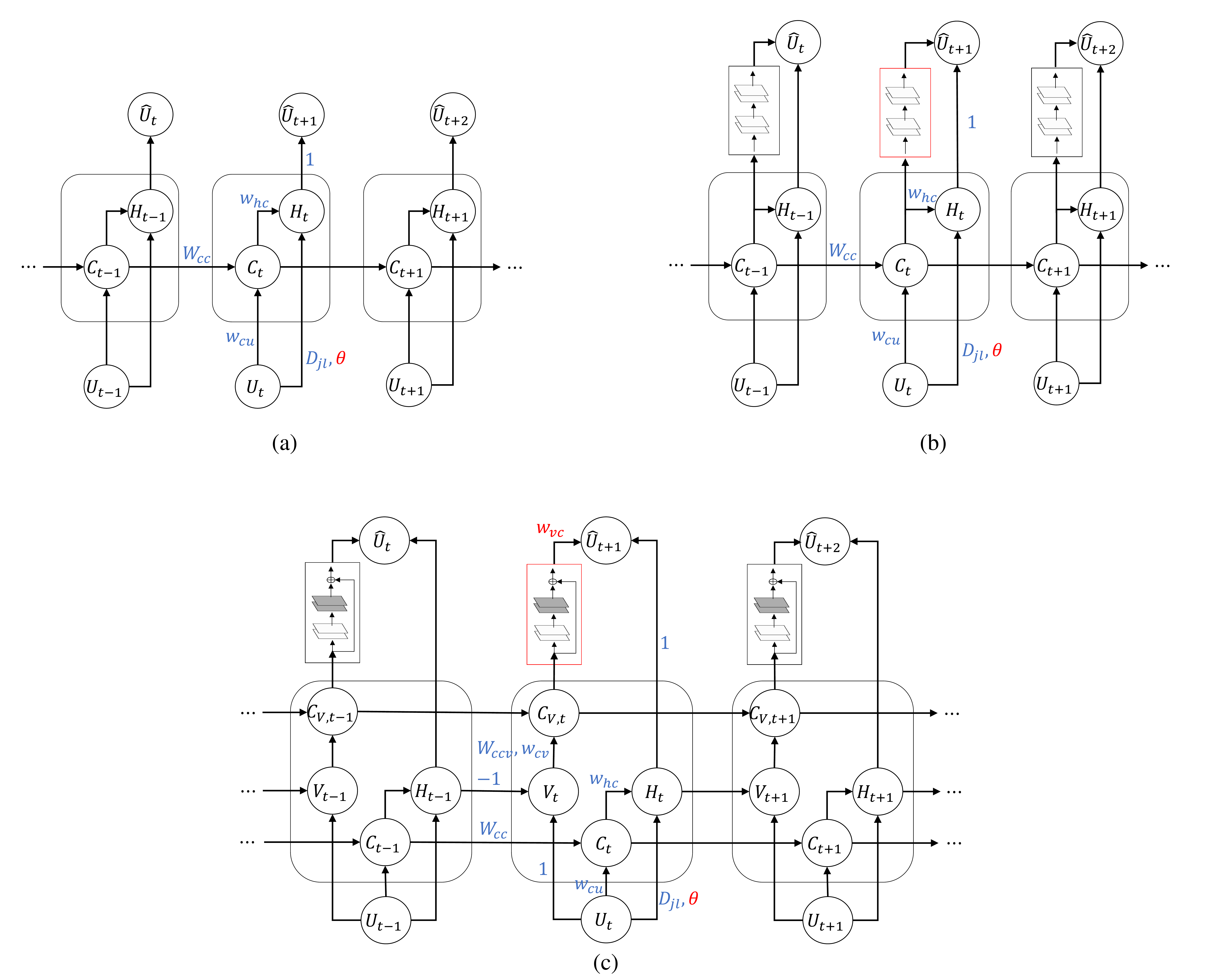}
  \caption{(a): PDE-RNN: RNN structure (unfolded) that maps a generic homogeneous PDE. $U_t$ is the observed map at time step $t$ and $\widehat{U}_{t+1}$ is the prediction of the next map. $H_t$ and $C_t$ are internal states of the RNN representing the homogeneous solution and cell memory, respectively. For homogeneous PDE, $\widehat{U}_{t+1} = H_t$. (b): PDE-RNN + CNN: a corrective convolutional module is added with the homogeneous solution to incorporate the source term or perturbation. 
  (c): PhICNet: Proposed RNN structure that models the dynamical system with time-varying independent source. A residual encoder-decoder network, which models the source dynamics, is integrated with (a). $V_t$ stores the estimated perturbation. $C_{V, t}$ is the cell memory for storing past estimated perturbation maps. In all three models, parameters of the connections are highlighted either in blue or red. The blue-colored parameters are fixed, whereas the red-colored parameters and the parameters inside red boxes are trainable.}
  \label{fig:model}
\end{figure*}

\begin{figure*}
  \centering
  \includegraphics[width=0.9\linewidth]{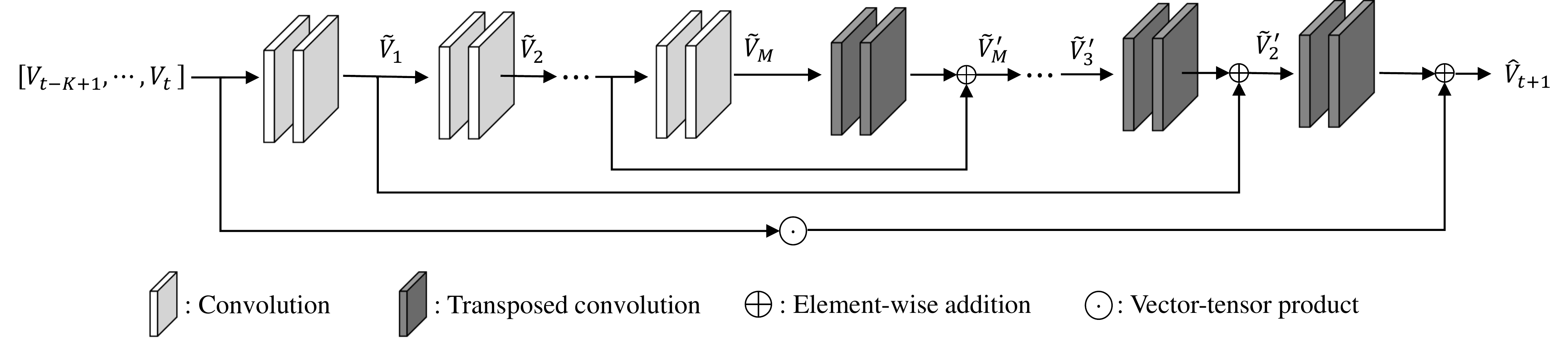}
  \caption{Architecture of the residual encoder-decoder network used for source dynamics modeling.}
  \label{fig:rednet}
\end{figure*}

\subsection{Background on Recurrent Neural Networks}

The recurrent neural network (RNN) is an elegant generalization of feedforward neural networks to incorporate temporal dynamics of data \cite{rumelhart1986learning, werbos1990backpropagation, schuster1997bidirectional}. The RNN and its various evolved topologies have proven efficacious in numerous sequence modelling tasks \cite{pearlmutter1989learning, robinson1994application, hochreiter1997long, sutskever2014sequence, xingjian2015convolutional}.
At each time step, an input vector $i_t$ is fed to the RNN. The RNN modifies its internal state $h_t$ based on the current input and previous internal state. The updated internal state is then used to predict the output $o_t$. The following set of equations (equation \ref{eqn:rnn}) delineates the computation inside a standard RNN.
\begin{align}
    h_t &= \sigma_h(W_{hi} i_t + W_{hh} h_{t-1} + b_h) \nonumber \\
    o_t &= \sigma_o(W_{oh} h_t + b_o) 
    \label{eqn:rnn}
\end{align}
$W_{hi}, W_{hh}, W_{oh}$ are the weight matrices of the RNN and $b_h, b_o$ are bias vectors. $\sigma_h$ and $\sigma_o$ are nonlinear activation functions. Temporal update in the internal state allows the RNN to make use of past information while predicting the current output.

The input $i_t$, internal state $h_t$ and output $o_t$ of standard RNN are all 1D vectors and the operations are fully-connected. 
To deal with 2D image data, Xingjian et al. proposed convolutional LSTM  \cite{xingjian2015convolutional} that uses convolutional operations instead of fully-connected operations of standard LSTM \cite{hochreiter1997long}, an evolved variant of the RNN. Incorporating convolutional operations in RNN, we can write the computation inside a convolutional-RNN as follows:
\begin{align}
    H_t &= \sigma_h (W_{hi} * I_t + W_{hh} * H_{t-1} + b_h) \nonumber \\
    O_t &= \sigma_o (W_{oh} * H_t + b_o)
\end{align}
where, $I_t$, $H_t$ and $O_t$ are the input, internal state and output, respectively, of the convolutional-RNN at time step $t$ and are all 2D images. `*' denotes the convolution operator. 

\subsection{Proposed RNN Model for Generic Homogeneous PDE}

Figure \ref{fig:model}(a) illustrates the structure of the RNN we propose for modeling a generic homogeneous PDE (i.e. zero source term); we will refer the model as PDE-RNN. Inputs to the RNN cell are 2D observation maps $U_t \in \mathbb{R}^{X \times Y}, t \in \{0,\ldots,T\}$. The RNN cell keeps an memory that stores the past information required for current step prediction. The concept of  cell memory was introduced in LSTM \cite{hochreiter1997long}. Cell memory in LSTM is controlled by several self-parameterized gates to learn what information to store and what information to forget. In contrast, past information required to be stored in the cell memory in our physics-incorporated RNN is completely determined beforehand based on the known temporal order $n$ (in equation \ref{eqn:main_dyn}) of the observed system. For an $n^{th}$ order (temporal) system, the cell memory stores the current and past $(n-1)$ observed maps. At time step $t$, the state of the cell memory (we will call it cell state from now on) $C_t$ defined by the following equation
\begin{equation}
    C_t = [U_t , \ldots, U_{t-n+1}]
    \label{eqn:ct_concat}
\end{equation}
where $[\cdot]$ denotes the concatenation operation along a new dimension. The cell state can be seen as a 3D tensor ($C_t \in \mathbb{R}^{n \times X \times Y})$. Cell state at current time step $t$ (equation \ref{eqn:ct_concat}) can be rewritten as a function of previous cell state $C_{t-1}$ and current input $U_t$:
\begin{equation}
    C_t = W_{cc} \  \circledcirc C_{t-1} \  + \  w_{cu} \circ U_t 
    \label{eqn:ct_update}
\end{equation}
$W_{cc}$ is a 2D square matrix of order $n$ and the $(p,q)^{th}$ element of $W_{cc}$, $p \in \{1, \ldots, n\}$
and $q \in \{1, \ldots, n\}$, is defined as follows.
\begin{equation}
    W_{cc}^{pq} = \begin{cases}
                  1 & \textit{if \  $p = q + 1$} \\
                  0 & \textit{otherwise}
                  \end{cases}
\end{equation}
`$\circledcirc$' denotes a matrix-tensor product resulting in a tensor $\Tilde{C} \in \mathbb{R}^{n \times X \times Y}$ such that
\begin{equation}
    \Tilde{C}^p = \sum_{q=1}^{n} W_{cc}^{pq} C_{t-1}^q, \quad p \in \{1, \ldots, n\}.
\end{equation}
Note, $C_{t-1}^q$ denotes the $q^{th}$ element of the tensor $C_{t-1} \in \mathbb{R}^{n \times X \times Y}$, where $q \in \{1, \ldots, n\}$. Therefore, according to equation \ref{eqn:ct_concat}, $C_{t-1}^q = U_{t-q}, \  q \in \{1, \ldots, n\}$.

The operator $\circ$ between 2D observation map $U_t \in \mathbb{R}^{X \times Y}$ and 1D vector $w_{cu} = [1, 0, \ldots, 0]^T \in \mathbb{R}^n$ performs a vector-matrix product to yield a tensor $\Acute{C} \in \mathbb{R}^{n \times X \times Y}$ such that
\begin{equation}
    \Acute{C}^p = w_{cu}^p U_t, \quad  p \in \{1, \ldots, n\}.
\end{equation}
Cell state $C_t$ and input $U_t$ are used to compute $H_t \in \mathbb{R}^{X \times Y}$ as follows.
\begin{align}
    H_t =& \ \   w_{hc} \odot C_t \nonumber \\
        &+ f(U_t, D_{10} * U_t, D_{01} * U_t, D_{11} * U_t, \ldots; \  \theta)
    \label{eqn:Ht}    
\end{align}
$H_t$ represents the solution of the system that is governed by a homogenous PDE. In other words, $H_t$ corresponds to the predicted map at timestep $t$ when there is no source term or perturbation. Equation \ref{eqn:Ht} is the discretized (finite difference) implementation of the homogeneous PDE (equation \ref{eqn:main_dyn} without the source term $v(x, y, t)$). 
$w_{hc} \in \mathbb{R}^n$ is determined by the temporal order of the dynamics. The elements of $w_{hc}$ are the coefficients of past observation maps in the finite difference approximation of $\frac{\partial^n}{\partial t^n}$ and are given by the following equation.
\begin{equation}
    w_{hc}^{p} = (-1)^{p+1} \binom{n}{p}, \quad p \in \{1, \ldots, n\}
\end{equation}
`$\odot$' denotes a vector-tensor product resulting in a 2D matrix $\Tilde{H} \in \mathbb{R}^{X \times Y}$ such that
\begin{equation}
    \Tilde{H} = \sum_{p=1}^{n} w_{hc}^{p} C_{t}^p
    \label{eqn:vtp}
\end{equation}

Function $f$ (in equation \ref{eqn:Ht}) is the implementation of $F$ (in equation \ref{eqn:main_dyn}) for discretized observation maps. As mentioned in section \ref{sec:problem}, the analytical form of $F$ or $f$ is known to us, but the physical parameters $\theta$ are unknown and trainable. Spatial derivatives of observation map $U_t$ are computed as convolution with differential kernels. $D_{jl}$ denotes the differential kernel corresponds to $\frac{\partial^{j+l}}{\partial x^j \partial y^l}$. Therefore, the convolution operation $D_{jl} * U_t$ in equation \ref{eqn:Ht} represents $\frac{\partial^{j+l}u}{\partial x^j \partial y^l}$ of the equation \ref{eqn:main_dyn}. The size and elements of a differential kernel are determined by the finite difference approximation of corresponding derivative. 

\begin{figure}
  \centering
  \includegraphics[width=1\linewidth]{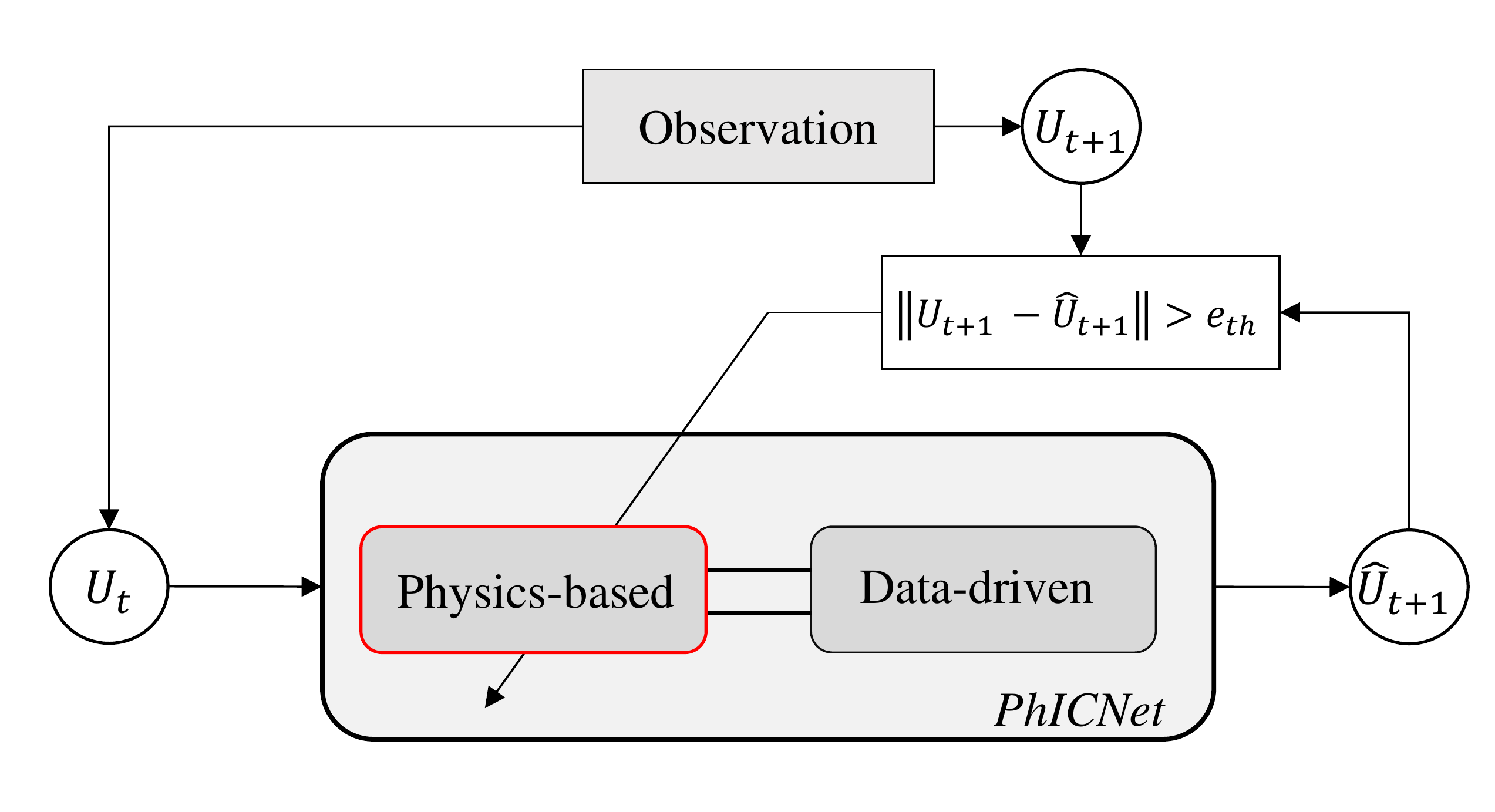}
  \caption{Online learning process of physical parameters. Only the parameters of the physical model (shown in red box) are re-learned when error between prediction and observation crosses a threshold $e_{th}$; parameters of the data-driven model (i.e. RED-Net) are kept frozen.}
  \label{fig:online_proc}
\end{figure}

\subsection{Approach to Incorporate Source Dynamics}

The basic structure of RNN for homogeneous PDE (PDE-RNN) needs to be modified to incorporate dynamic source term. Note again, the source term cannot be learned separately as it cannot be observed directly. Therefore, the RNN should be structured in such a way that it inherently identifies the source dynamics. 
A simple modification can be adding a convolutional neural network (CNN) with the PDE-RNN. We will call this modified structure (Figure \ref{fig:model}(b)) PDE-RNN + CNN. The convolutional network takes the cell state $C_t$ as input and add a corrective term, which accounts for the source or perturbation, to the homogeneous solution. Although PDE-RNN + CNN incorporates the known PDE in its structure, it does not get much benefit in forecasting the system or identifying the source dynamics as we will show in section \ref{sec:experiments}. 

In contrast to directly predicting a correcting term from observation maps like PDE-RNN + CNN, we propose to estimate an intermediate source map and use that to learn the source dynamics. Separating the source maps from the overall dynamics makes it easier to learn the source dynamics. Figure \ref{fig:model}(c) shows the proposed PhICNet structure. An internal state $V_t$ is added in the cell that estimates the perturbation using the predicted homogeneous solution from the previous step and the current input. $V_t \in \mathbb{R}^{X \times Y}$ is computed as follows.
\begin{equation}
    V_t = U_t \ - \  H_{t-1}
\end{equation}
Another cell memory $C_{V, t}$ is added to store past perturbation estimates. Assuming a $K^{th}$ order source dynamics, $C_{V, t}$ can be seen as a 3D tensor ($C_{V, t} \in \mathbb{R}^{K \times X \times Y})$) given by:
\begin{equation}
    C_{V, t} = [V_t , \ldots, V_{t-K+1}]
    \label{eqn:cvt}
\end{equation}
Cell memory $C_{V, t}$ is updated at each time step by the following equation.
\begin{equation}
    C_{V, t} = W_{ccv} \  \circledcirc C_{V, t-1} \  + \  w_{cv} \circ V_t 
\end{equation}
$W_{ccv} \in \mathbb{R}^{K \times K}$ and $w_{cv} \in \mathbb{R}^K$ have similar properties of $W_{cc}$ and $w_{cu}$, respectively, of equivalent equation \ref{eqn:ct_update}. 
Finally, the predicted map $\widehat{U}_{t+1}$ is computed by the following equation.
\begin{equation}
    \widehat{U}_{t+1} = H_t \ + \ w_{vc} \odot C_{V,t} + \  g(C_{V,t})
    \label{eqn:Ut+1}
\end{equation}
Function $g$, the implementation of fuction $G$ (in equation \ref{eqn:source_dyn}) for discretized source maps, captures the source dynamics. We use a residual convolutional network for this purpose such that
\begin{equation}
    \widehat{V}_{t+1} = \ w_{vc} \odot C_{V,t} + \  g(C_{V,t})
\end{equation}
$w_{vc} \odot C_{V,t}$ is a vector-tensor product similar to equation \ref{eqn:vtp}. The vector $w_{vc} \in \mathbb{R}^K$ is a trainable parameter and initialized with coefficients of the finite difference approximation of $\frac{\partial^K}{\partial t^K}$. $\widehat{V}_{t+1}$ is the predicted source map  which is added to the homogeneous solution $H_t$ to get the predicted map $\widehat{U}_{t+1}$ (equation \ref{eqn:Ut+1}). 

\begin{figure*}
  \centering
  \includegraphics[width=0.8\linewidth]{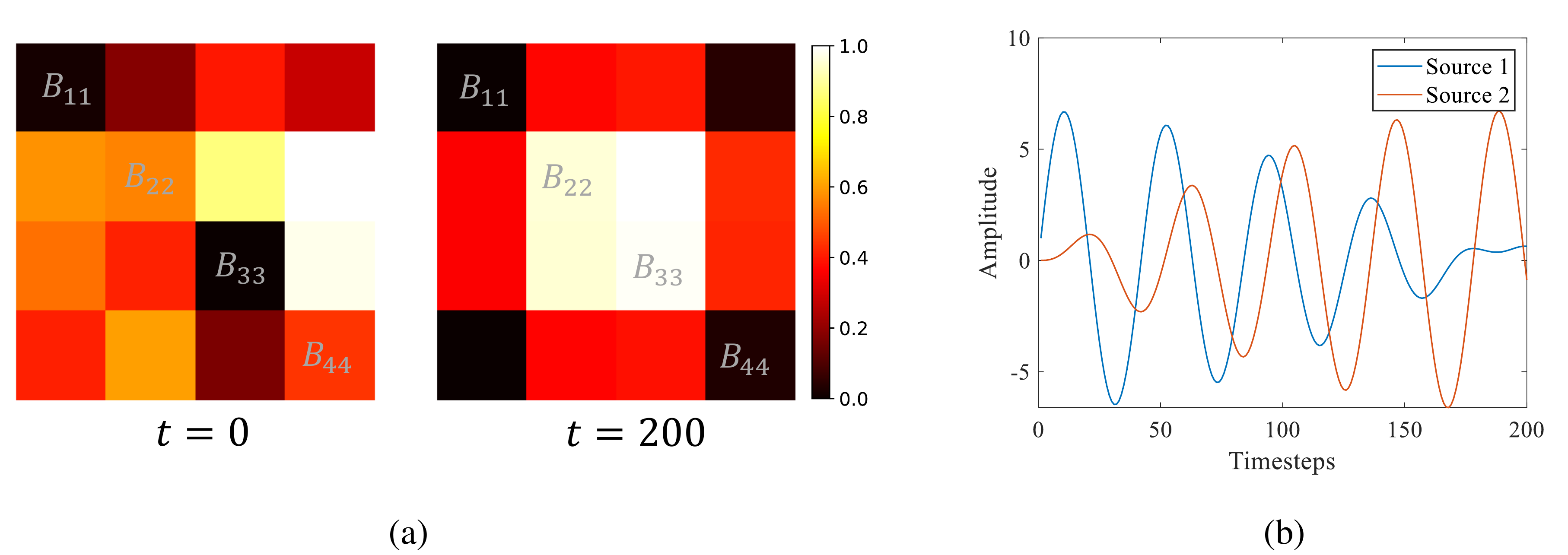}
  \caption{(a): Heat-source maps at the initial and final time step of an example sequence used in our experiment. (b): Temporal behavior of the coupled oscillators acting as sources in an example sequence used in our wave propagation experiment.}
  \label{fig:sources}
\end{figure*}

\begin{figure*}
  \centering
  \includegraphics[width=0.98\linewidth]{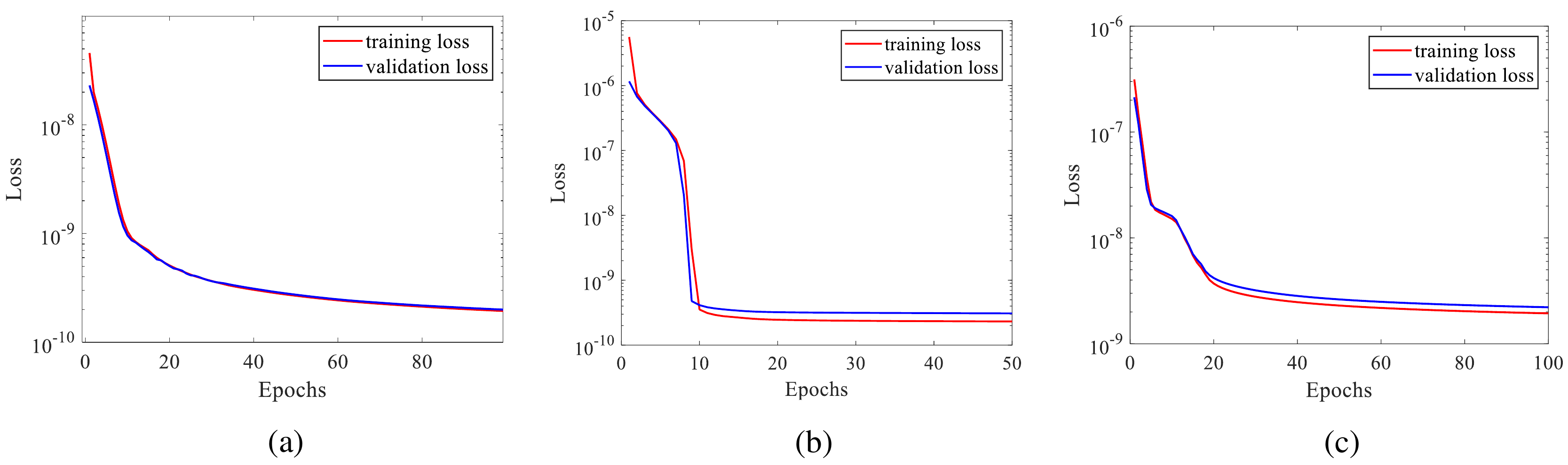}
  \caption{Training loss and validation loss of PhICNet over epochs for heat diffusion system (a), wave propagation system (b) and Burgers’ fluid flow system (c).}
  \label{fig:losses}
\end{figure*}

The motivation behind using residual network for modeling the source dynamics is that several studies have established the connection between residual networks and differential equations \cite{weinan2017proposal,lu2018beyond,chang2018reversible,chen2018neural}. We use an architecture similar to residual encoder-decoder network (RED-Net) \cite{mao2016image}. To incorporate $K^{th}$ order source dynamics, we use past $K$ estimated source maps stacked across the channel dimension as input to RED-Net. Moreover, we add a weighted combination of past $K$ estimated source maps $w_{vc} \odot C_{V,t}$, instead of adding just $V_t$, to the final output. 
The elements of $w_{vc}$ are initialized with the coefficients of the finite difference approximation of $\frac{\partial^K}{\partial t^K}$.

Figure \ref{fig:rednet} shows the architecture of the residual encoder-decoder network with $M$ convolutional and $M$ transposed convolutional blocks. Each convoltutional block consists of two convolutional layers. Similarly, each transposed convolutional block comprises two transposed convolutional layers. Convolutional encoder extracts feature at different scales. These feature maps are used by the transposed convolutional decoder with symmetric skip connections from corresponding convolutional block to capture dynamics at different scales. Skip connection also allows to use deeper network for complex dynamics without encountering the problem of vanishing gradient. 

The computation at the $m^{th}$ convoltional block is given by the following equation. 
\begin{align}
    \Tilde{V}_m = \sigma \big(W_{m2} \  * \ \sigma(W_{m1} \ * \ \Tilde{V}_{m-1})\big), \nonumber \\ m \in \{1,  \ldots, M\}, \nonumber \\
    \Tilde{V}_0 = C_{V,t} 
\end{align}
The computation at the $m^{th}$ transposed convolution block from the end is delineated by  
\begin{align}
    \Tilde{V}_m^{'} = \sigma \big(W_{m2}^{'} \  \star \ \sigma(W_{m1}^{'} \ \star  \ \Tilde{V}^{'}_{m+1}) \ + \ \Tilde{V}_{m-1} \big), \nonumber \\
    m \in \{2, \ldots, M\}, \nonumber \\
    \Tilde{V}^{'}_{M+1} =  \Tilde{V}_M
\end{align} and,
\begin{align}
    \Tilde{V}_1^{'} = W_{12}^{'} \  \star \ \sigma(W_{11}^{'} \ \star  \ \Tilde{V}^{'}_{2}) \ + \ \ w_{vc} \odot \Tilde{V}_{0}
\end{align}

The predicted source map is given by $\widehat{V}_{t+1} = \Tilde{V}_1^{'} $. $\star$ denotes the transposed convolution operation and $\sigma$ is the activation function ReLU. 

\subsection{Training Loss}
For a sequence of observation maps $\{U_0, U_1, \ldots, U_T\}$ and $n^{th}$ order (temporal) system, assuming a $K^{th}$ order source dynamics, the prediction loss is defined as follows.
\begin{equation}
    \mathcal{L}_{pred} = \frac{1}{T-n-K+1} \sum_{t=n+K}^{T} \| U_{t} - \widehat{U}_{t} \|_2^2
\end{equation}
Estimated source map $V_t$, after observing $U_t$ at timestep $t$, should match with predicted source map $\widehat{V}_{t}$ from previous timestep. Accordingly, we add a source prediction loss to the training objective, given by 
\begin{equation}
    \mathcal{L}_{source\underline{\hspace{0.1cm}}pred} = \frac{1}{T-n-K+1} \sum_{t=n+K}^{T} \| V_{t} - \widehat{V}_{t} \|_2^2
\end{equation}
Furthermore, source map can be densely distributed or sparse (may contain only a single source). To deal with source map sparsity, we add a $L1$ penalty :
\begin{equation}
    \mathcal{L}_{source\underline{\hspace{0.1cm}}sparse} = \frac{1}{T-n-K+1} \sum_{t=n+K}^{T} \|\widehat{V}_{t} \|_1
\end{equation}
The overall loss for training is
$\mathcal{L} = 
        \mathcal{L}_{pred} +
        \mathcal{L}_{source\underline{\hspace{0.1cm}}pred} +
        \lambda \mathcal{L}_{source\underline{\hspace{0.1cm}}sparse}$, 
where $\lambda$ is a hyperparameter. 

\subsection{Online Learning of Time-varying Physical Parameters}
\label{subsec:online}
The aforementioned proposed method learns the parameters of a physical model in conjunction with a data-driven model for unknown source dynamics. The model is trained with dataset comprising a fixed set of physical parameters. However, physical parameters of real-world physical systems are often not fixed and can change over time. Adapting the parameters of a (pure) physical model with online observation is well-established in literature. Here, we would like to verify if adaptation of physical parameters with online observation is feasible for hybrid models like ours. We take a model which is trained with a fixed value of the physical parameter and employ it to predict the dynamics of a system where the value of the physical parameter changes over time. We assume that the true observation $U_t$ is available online to compare against the predicted map $\widehat{U}_t$. At each time step, we compare the prediction with (true) observed state; if the error is higher than a threshold, the parameters of the physical model are re-tuned using gradient descent to reduce the error while parameters of the data-driven model (i.e. RED-Net) are kept frozen. Note, we consider the case where only the physical parameters are varying but the source dynamics remains same.
The process is outlined in Figure \ref{fig:online_proc}.

%% file: experiments.tex
\section{Experimental Evaluation}
\label{sec:experiments}

\begin{figure*}
  \centering
  \includegraphics[width=0.95\linewidth]{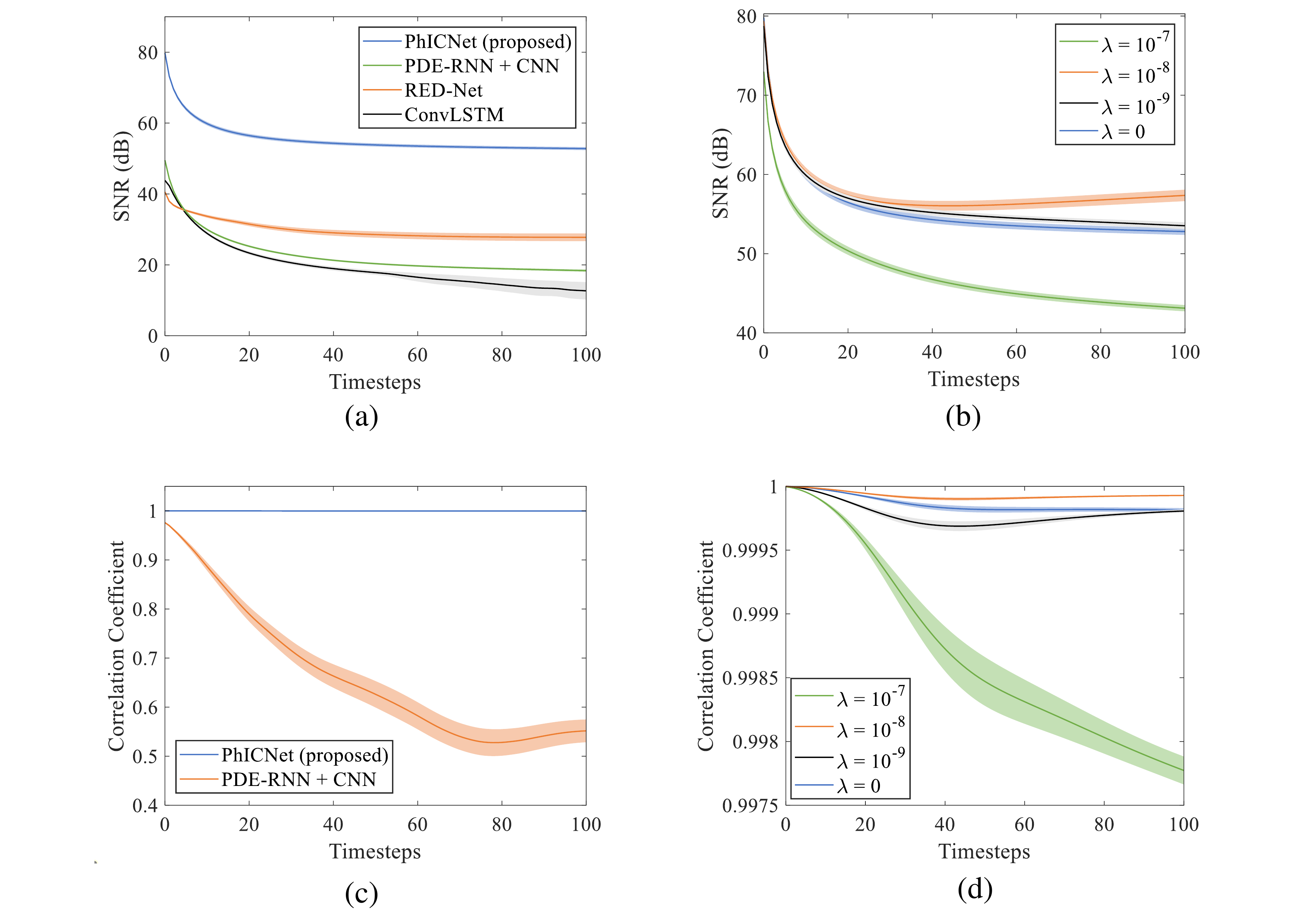}
  \caption{Quantitative analysis for the \textbf{heat diffusion system}. (a, c):  Accuracy comparison of proposed PhICNet with respect to other baselines in the task of forecasting (a) and source identification (c). (b, d): Effect of sparsity hyperparameter $\lambda$ on forecasting (b) and source identification (d) performance of the proposed model. In all plots, shaded areas show $95\%$ confidence interval.}
  \label{fig:heat_metrics}
\end{figure*}

\begin{figure*}
  \centering
  \includegraphics[width=1\linewidth]{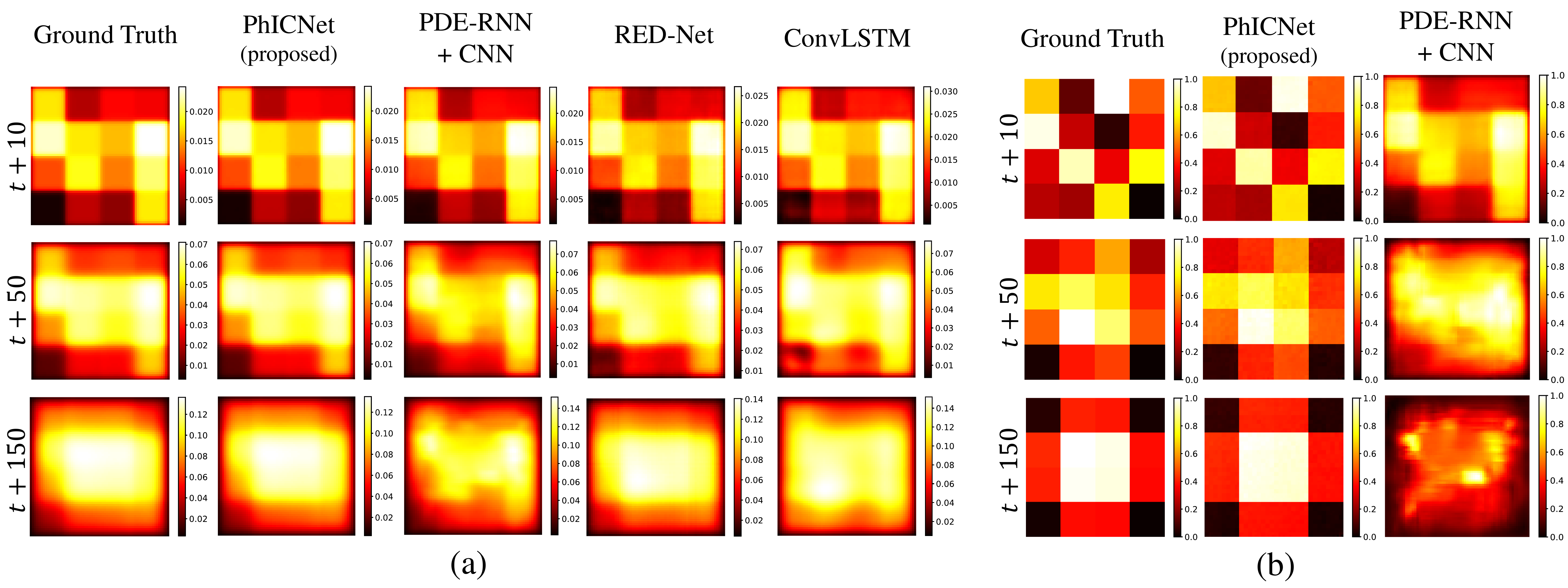}
  \caption{
  Qualitative analysis for the \textbf{heat diffusion system}. Visual comparison of predicted heat maps (a) and source maps (b) by different models at time steps $t + 10$, $t + 50$ and $t + 150$ when last observation is taken at $t$. Colorbar on the right side of each colormap shows its numerical range.}
  \label{fig:heat_visual}
\end{figure*}

We evaluate our model on three dynamical systems: heat diffusion system, wave propagation system and Burgers’ fluid flow system. Heat diffusion system and Burgers’ fluid flow system have temporal order of 1 ($n=1$), while wave propagation system is a second order system ($n=2$). For the task of forecasting of the overall dynamics, we compare the proposed model with PDE-RNN + CNN, ConvLSTM \cite{xingjian2015convolutional}, a residual encoder-decoder network. However, among the baselines, only PDE-RNN + CNN can be used for source identification task. 

In contrast to the residual encoder-decoder used in our model to predict only the source dynamics, the baseline residual encoder-decoder network (RED-Net) models the combined dynamics. Accordingly, the input and output of the baseline residual encoder-decoder network are observation maps ($U$). For RED-Net baseline, we assume the temporal order of the system is known a priori, i.e. for an $n^{th}$ order system, assuming a $K^{th}$ order source dynamics, input to the model is the sequence $\{U_t, U_{t-1}, \ldots, U_{t-n-K+1}\}$ while predicting $\widehat{U}_{t+1}$. 

All the models are implemented in PyTorch framework. We run all the experiments on a computer equipped with Intel Core i7-8700K CPU (with 62GB RAM) and NVIDIA GTX 1080Ti GPU (with 12GB RAM). Average CPU inference time is $1.6$ms per time-step while average GPU inference time is $1.1$ms per time-step.

\begin{figure*}
  \centering
  \includegraphics[width=0.9\linewidth]{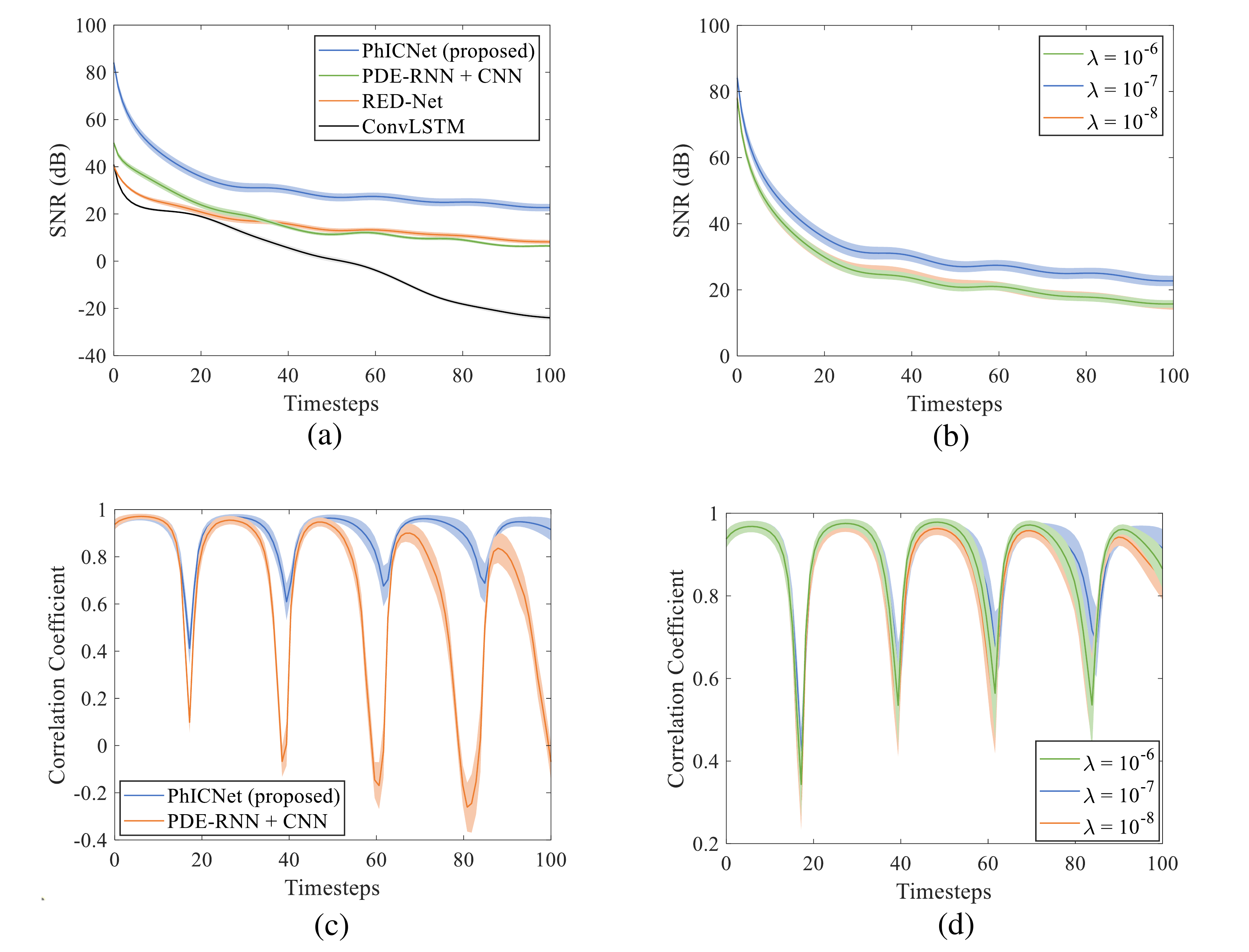}
  \caption{Quantitative analysis for the \textbf{wave propagation system}. (a, c):  Accuracy comparison of proposed PhICNet with respect to other baselines in the task of forecasting (a) and source identification (c). (b, d): Effect of sparsity hyperparameter $\lambda$ on forecasting (b) and source identification (d) performance of the proposed model. In all plots, shaded areas show $95\%$ confidence interval.}
  \label{fig:wave_metrics}
\end{figure*}

\begin{figure*}
  \centering
  \includegraphics[width=1\linewidth]{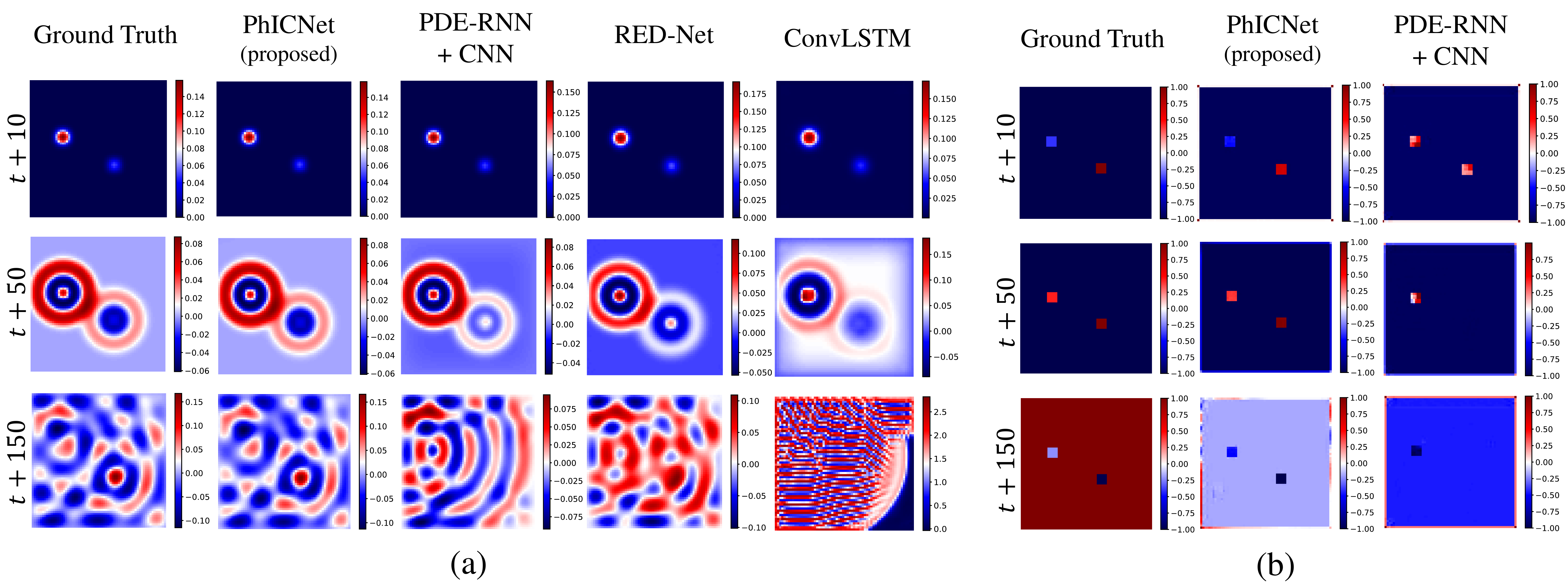}
  \caption{Qualitative analysis for the \textbf{wave propagation system}. Visual comparison of predicted wave maps (a) and source maps (b) by different models at time steps $t + 10$, $t + 50$ and $t + 150$ when last observation is taken at $t$. Colorbar on the right side of each colormap shows its numerical range.}
  \label{fig:wave_visual}
\end{figure*}

\begin{figure*}
  \centering
  \includegraphics[width=0.9\linewidth]{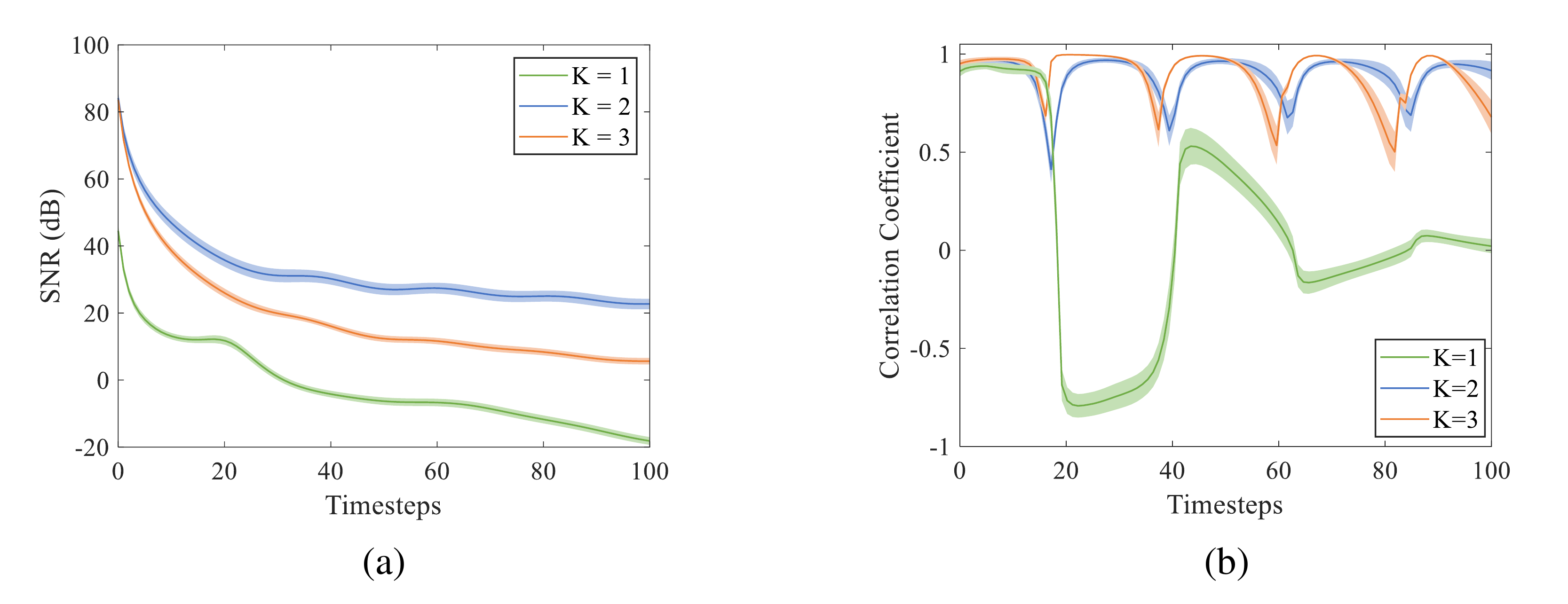}
  \caption{Choice of temporal order $K$ in source dynamics modeling: effect on forecasting (a) and source identification (b) performance for the \textbf{wave propagation system}.}
  \label{fig:order}
\end{figure*}

\subsection{Heat Diffusion System}
Heat diffusion at the surface of a material is described by:
\begin{align}
    \frac{\partial u}{\partial t} = \alpha \bigg(\frac{\partial^2 u}{\partial x^2} \  + \  \frac{\partial^2 u}{\partial y^2}\bigg) \ + \ v(x, y, t), \nonumber \\ 
    \qquad (x, y) \in \Omega \subset \mathbb{R}^2, t \in [0,T]
    \label{eqn:heat}
\end{align}
where $u(x,y,t)$ is the heat density at location $(x,y)$ at time $t$ and $v(x, y, t)$ is the perturbation due to heat source(s). $\alpha$ is the thermal diffusivity of the material, the rate at which heat disperses through that material. Equation \ref{eqn:heat} is one of the fundamental PDEs and is used to describe diffusion of heat, chemicals, brownian motion, diffusion models of population dynamics, and many other phenomena \cite{strauss2007partial}.

The computation space $\Omega$ is discretized into $64 \times 64$ regular mesh, i.e. $U_t \in \mathbb{R}^{64 \times 64}$. For heat-source, we consider the source map $V_t \in \mathbb{R}^{64 \times 64}$ is divided into 16 equal-sized blocks initialized with random values in $[0,1]$. All grid points belonging to a block $B_{jl}$ take same value at any time step. The evolution of source map happens in the block level. Each block $B_{jl}$ follows a dynamics given by:
\begin{equation}
    \frac{dV_t^{B_{jl}}}{dt} =  \sum_{(r,s) \in \mathcal{N}(j, l)} \gamma (V_t^{B_{rs}} - V_t^{B_{jl}})
\end{equation}
where $V_t^{B_{jl}}$ denotes the value of block $B_{jl}$ at timestep $t$, $\gamma$ is a positive constant and $\mathcal{N}(j,l)$ represents the 4-connected neighborhood of block $B_{jl}$. Figure \ref{fig:sources}(a) shows an example of source map at the initial and final time step.
Training and test dataset are generated using numerical solution method starting from initial condition $U_{t < 0} = 0$ and assuming homogeneous Dirichlet boundary condition. Each sequence comprises $200$ frames and the training set contains $100$ such sequences while the test set contains $50$. $20\%$ of the training set is used for validation. 
In this system, the trainable parameters are diffusivity $\alpha$ and the parameters of the residual encoder-decoder network used to model the source dynamics. Since we need second-order spatial derivatives (equation \ref{eqn:heat}), the minimum size of the corresponding differential kernels should be $3 \times 3$. Specifically, following two differential kernels are used to compute $H_t$ in equation \ref{eqn:Ht}.
\begin{equation}
    D_{20} = \begin{pmatrix}
             0 & 0 & 0 \\
             1 & -2 & 1 \\
             0 & 0 & 0
             \end{pmatrix}, \quad 
    D_{02} = \begin{pmatrix}
             0 & 1 & 0 \\
             0 & -2 & 0 \\
             0 & 1 & 0
             \end{pmatrix}     
   \label{eqn:diff_kernels}             
\end{equation}
Training loss and validation loss of our model for this system are shown in Figure \ref{fig:losses}(a). 

We use Signal-to-Noise Ratio ($SNR$), defined in equation \ref{eqn:snr}, to quantitatively compare the performance of different models in forecasting the overall dynamics. 
\begin{equation}
    SNR(U_t, \widehat{U}_t) = 20 \log_{10} \frac{ \| U_{t} \|_2}{ \| U_{t} - \widehat{U}_{t} \|_2} 
    \label{eqn:snr}
\end{equation}
However, SNR cannot be used as a metric for source map comparison. To compare two maps using SNR, both maps need to have values in very similar scale. Since the inputs to the models are normalized and the source maps are learned as intermediate states without any direct supervisory signal, the estimated maps do not match the scale of true source maps. To quantify the similarity between true and estimated source maps, we use Correlation Coefficient as metric. Correlation coefficient between estimated source map $V_t$ and true source map $V_{true, t}$ is given by 
\\ \\ \\
\begin{align}
    &\rho (V_{true, t}, V_t) \nonumber \\ &= \frac{\sum_{j,l} (V_{true, t}^{jl} - \bar V_{true, t}) (V_t^{jl} - \bar V_t)}{\big ( \sum_{j,l} (V_{true, t}^{jl} - \bar V_{true, t})^2 \big )^{1/2} \big ( \sum_{j,l} (V_{t}^{jl} - \bar V_{t})^2 \big )^{1/2}}
\end{align}
where $\bar V_t$ and $\bar V_{true, t}$ denote the mean values of $V_t$ and $V_{true, t}$ respectively. 

Figure \ref{fig:heat_metrics} shows the quantitative comparison of proposed method with respect to other baselines and choice of hyperparameter $\lambda$ in the task of forecasting and source identification for the heat diffusion system. Though the source maps of the heat system do not appear as sparse visually, some blocks in the source maps have values very close to zero (for example, $B_{11}$ and $B_{33}$ at $t=0$, and $B_{11}, B_{14}, B_{41}$ and $B_{44}$ at $t=200$ in Figure \ref{fig:sources}(a)). This can be a possible explanation for attaining the best accuracy at a nonzero $\lambda$ (Figure \ref{fig:heat_metrics}(b,d)). It is also noteworthy that  even if the SNR of the predicted heatmaps for different values of $\lambda$ are significantly different (Figure \ref{fig:heat_metrics}(b)), the changes in the correlation coefficient of the estimated source maps is very small for different values of $\lambda$ (Figure \ref{fig:heat_metrics}(d)). On the other hand, correlation coefficient of the source maps, estimated by the PDE-RNN+CNN method, drops significantly over time (Figure \ref{fig:heat_metrics}(c)). 
Qualitative comparison of predicted heat maps by different models along with ground truth is depicted in Figure \ref{fig:heat_visual}(a).
PhICNet outperforms all the baselines. RED-Net is the best performing baseline. Effective modeling of dynamics by RED-Net is a key factor in the performance of our model as well since we use it for source dynamics modeling.
Source maps predicted by the proposed model and PDE-RNN + CNN are compared with ground truth in Figure \ref{fig:heat_visual}(b).

\subsection{Wave Propagation System}

Undulation in a stretched elastic membrane due to some perturbation can be described by:
\begin{align}
    \frac{\partial^2 u}{\partial t^2} = c^2 \bigg(\frac{\partial^2 u}{\partial x^2} \  + \  \frac{\partial^2 u}{\partial y^2}\bigg) \ + \ v(x, y, t), \nonumber \\ 
    \qquad (x, y) \in \Omega \subset \mathbb{R}^2, t \in [0,T]
    \label{eqn:wave}
\end{align}
where $u(x,y,t)$ is the deflection at location $(x,y)$ at time $t$ and $v(x, y, t)$ is the external perturbation. $c$ is the wave propagation speed.
We consider two coupled oscillators at random locations as wave sources perturbing the membrane. Figure \ref{fig:sources}(b) shows an example of temporal behavior of the sources. 

Similar to heat diffusion system, the computation space $\Omega$ is discretized into $64 \times 64$ regular mesh, i.e. $U_t \in \mathbb{R}^{64 \times 64}$. Unlike the source map considered for the heat system, the source map $V_t \in \mathbb{R}^{64 \times 64}$ for this wave system is sparse as the perturbation is applied only at two small regions of the membrane.  
The initial amplitude and location of the oscillators are chosen randomly for each sequence in the dataset. Therefore, source identification task requires identifying both the location and strength of the sources. Training and test dataset are generated using numerical solution method starting from initial condition $U_{t < 0} = 0$ and assuming homogeneous Dirichlet boundary condition. Each sequence comprises $200$ frames and the training set contains $300$ such sequences while the test set contains $50$. $20\%$ of the training set is used for validation. Trainable parameters for this system include propagation speed $c$ and the parameters of residual encoder-decoder network that is used to model the source dynamics. The differential kernels used for this system are same as equation \ref{eqn:diff_kernels}. 

\begin{figure}
  \centering
  \includegraphics[width=1\linewidth]{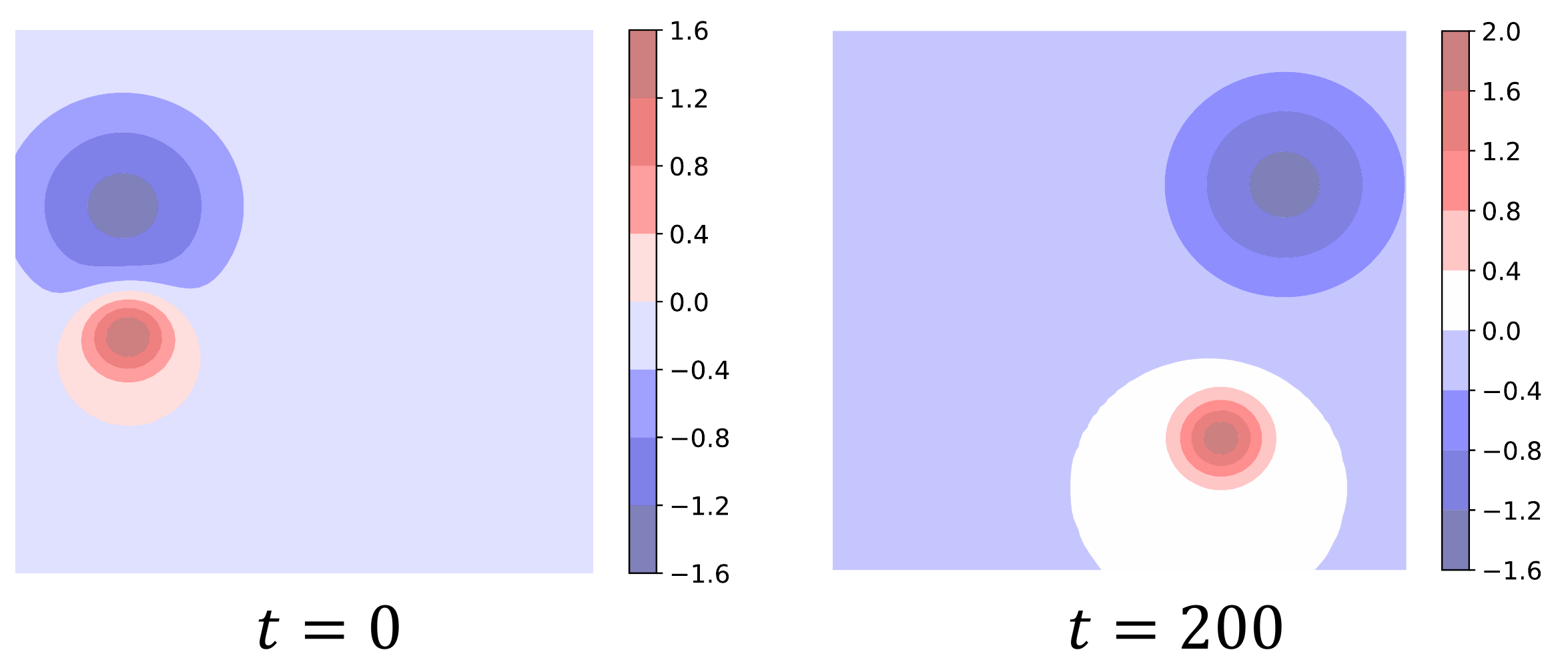}
  \caption{Pressure maps at the initial and final time step of an example sequence used in our \textbf{Burgers' fluid flow} experiment.}
  \label{fig:pressure_field}
\end{figure}

Training loss and validation loss of our model for this system are shown in Figure \ref{fig:losses}(b). Figure \ref{fig:wave_metrics} shows the quantitative comparison of proposed method with respect to other baselines and choice of hyperparameter $\lambda$ in the task of forecasting and source identification for the wave propagation system. A strong periodicity can be observed in the correlation coefficient plots (Figure \ref{fig:wave_metrics}(c,d)). The frequency of the valleys in correlation is related to the frequency of the source oscillators. The valleys occur when the oscillator of higher amplitude change its sign. The model predicts this sign change either sooner or later than when it actually happens and creates the valleys. However, these events cause only some insignificant fluctuation in SNR.
Qualitative comparison of predicted maps by different models along with ground truth is depicted in Figure \ref{fig:wave_visual}(a). Figure \ref{fig:wave_visual}(b) compares the source maps predicted by the proposed model and PDE-RNN + CNN with ground truth. \\ \\
\textbf{Choice of temporal order $K$ in source dynamics modeling.} Figure \ref{fig:order} compares the forecasting and source identification performance for different values of $K$ (in equation \ref{eqn:cvt}). The true order of the source dynamics (coupled oscillators) of the wave propagation system is 2. Best performance is observed when the source order is exactly known, i.e. $K=2$. Choosing an order higher than the true value ($K=3$) is better compared to choosing an order lower than the true value ($K=1$).

\begin{figure*}
  \centering
  \includegraphics[width=0.95\linewidth]{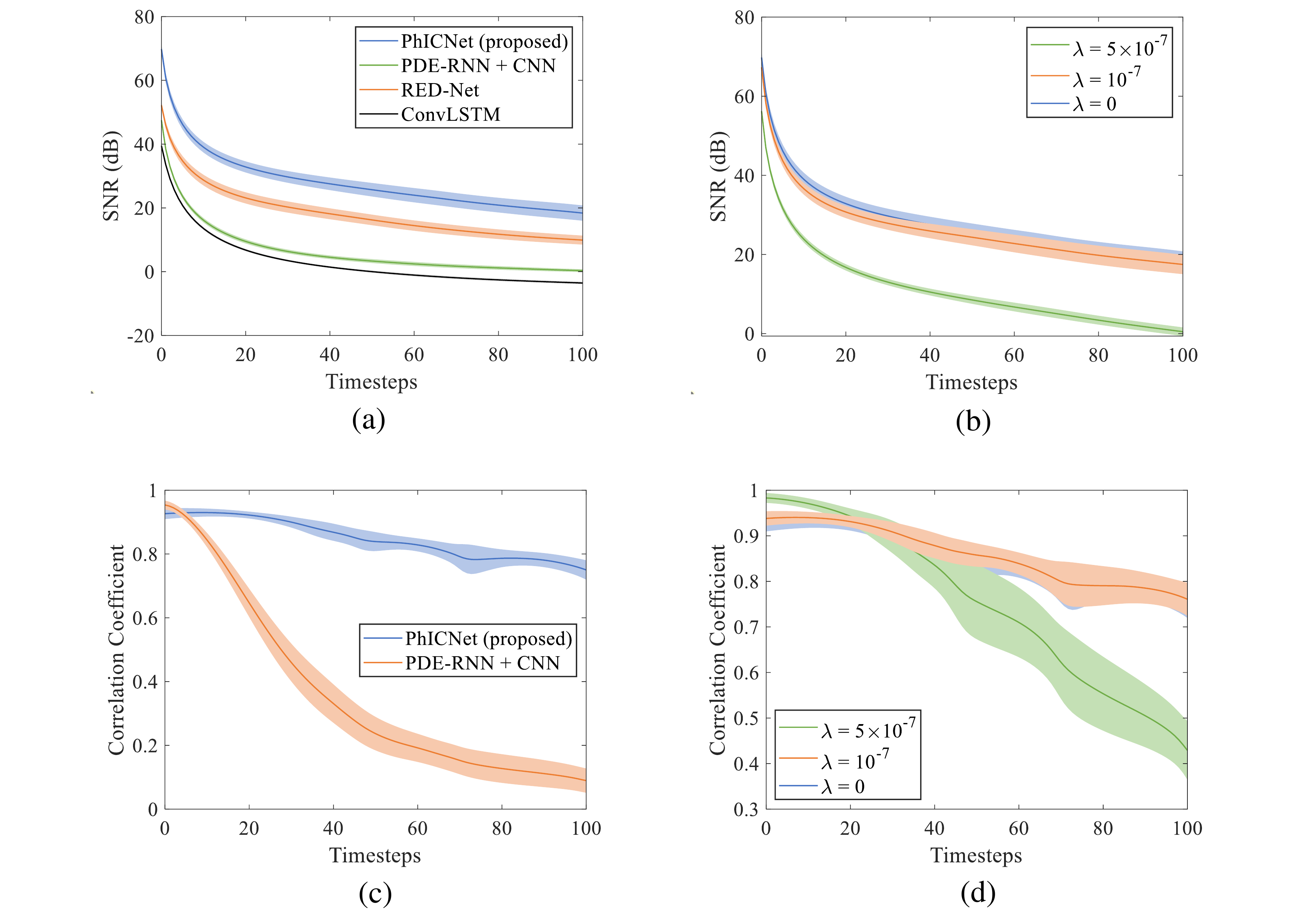}
  \caption{Quantitative analysis for the \textbf{Burgers' fluid flow system}. (a, c): Accuracy comparison of proposed PhICNet with respect to other baselines in the task of forecasting (a) and source identification (c). (b, d): Effect of sparsity hyperparameter $\lambda$ on forecasting (b) and source identification (d) performance of the proposed model. In all plots, shaded areas show $95\%$ confidence interval.}
  \label{fig:be_metrics}
\end{figure*}

\begin{figure*}
  \centering
  \includegraphics[width=1\linewidth]{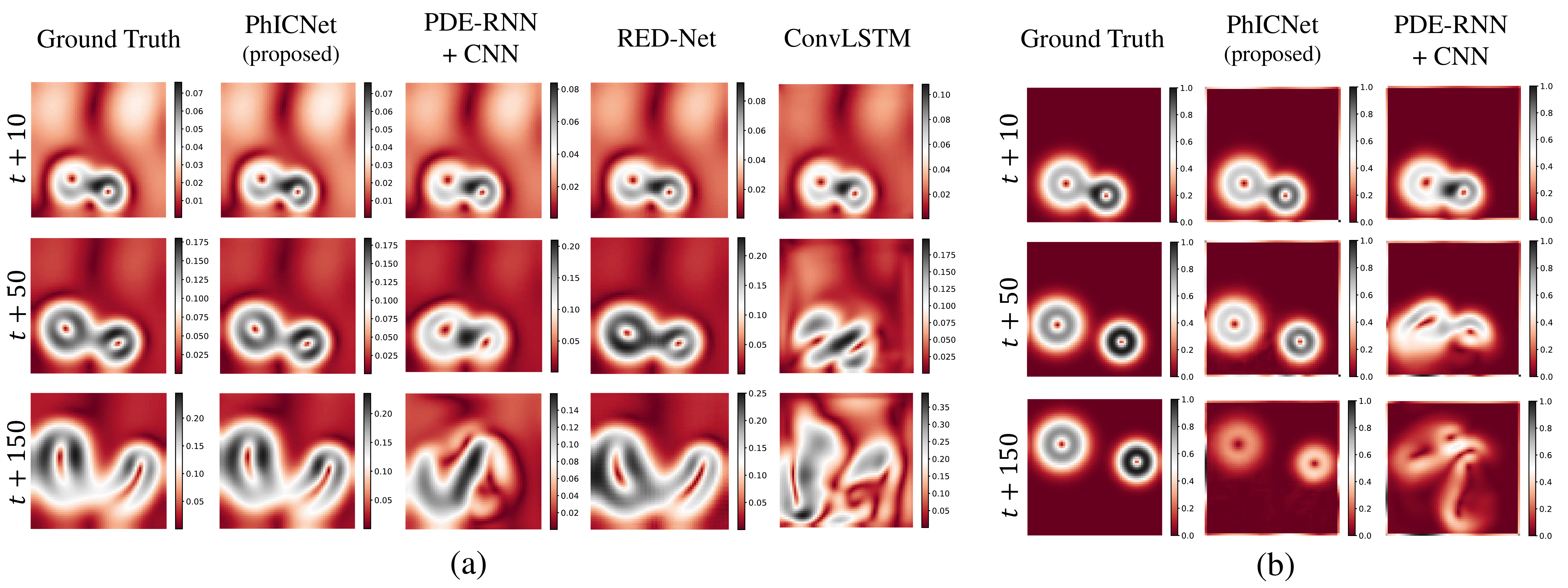}
  \caption{Qualitative analysis for the \textbf{Burgers' fluid flow system}. Visual comparison of predicted flow fields (a) and external pressure gradients (b) by different models at time steps $t + 10$, $t + 50$ and $t + 150$ when last observation is taken at $t$. The colormaps show the magnitude of flow velocity. Colorbar on the right side of each colormap shows its numerical range.}
  \label{fig:be_visual}
\end{figure*}

\begin{figure}
  \centering
  \includegraphics[width=0.85\linewidth]{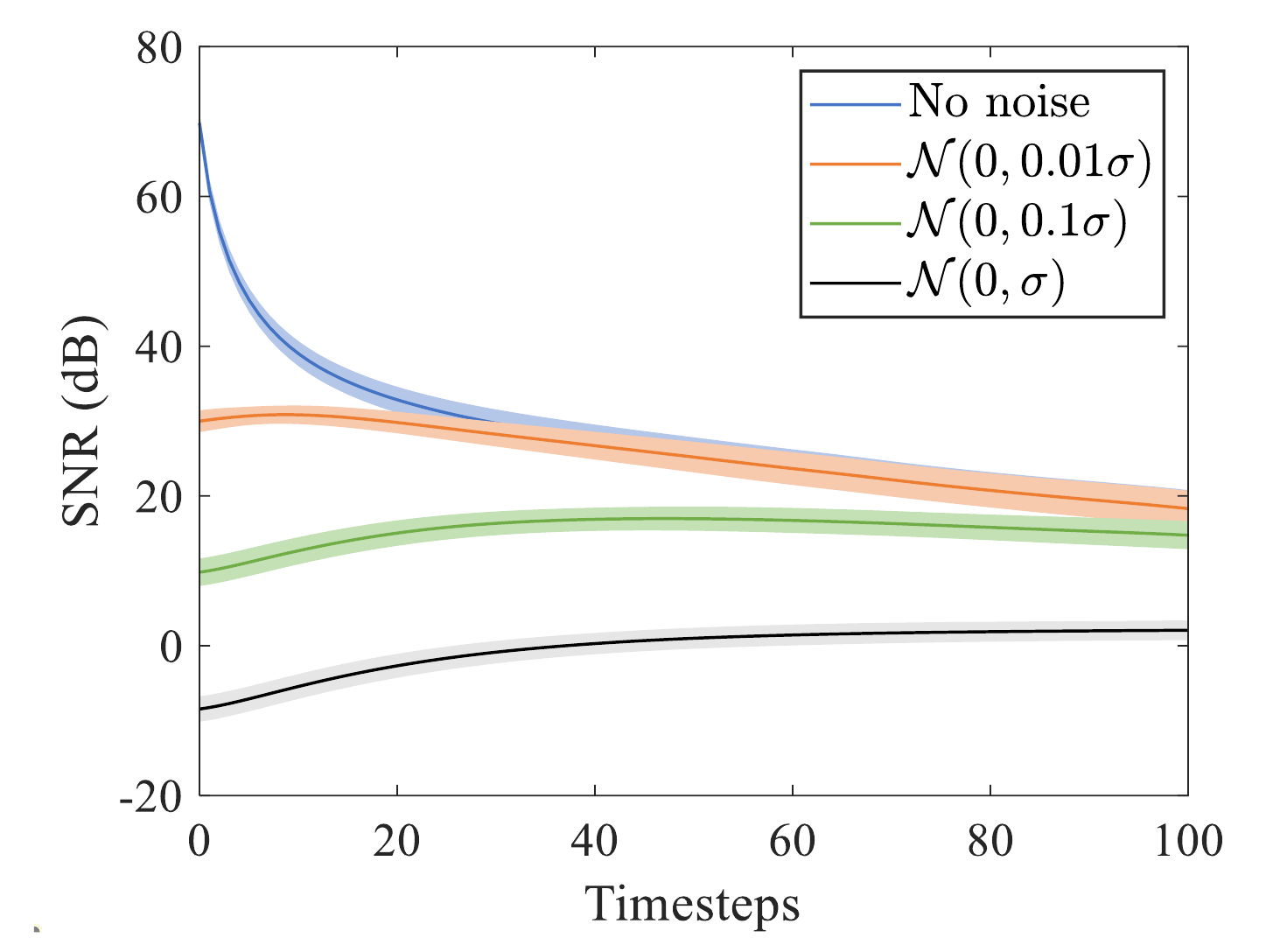}
  \caption{Performance of PhICNet in the forecasting task for \textbf{Burgers' fluid flow system} when the observation are corrupted with Gaussian noise of different variances. $\sigma$ is the standard deviation of the dataset.}
  \label{fig:noise}
\end{figure}

\begin{figure*}
  \centering
  \includegraphics[width=0.85\linewidth]{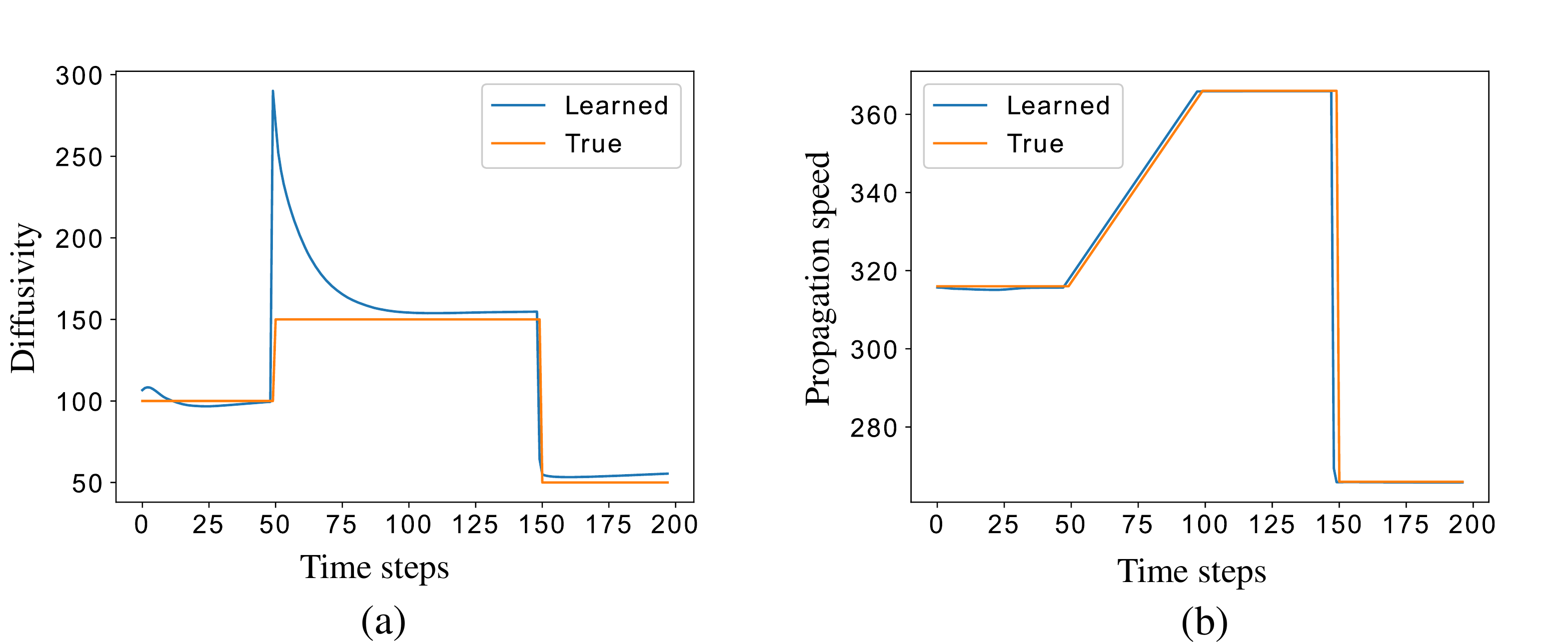}
  \caption{Online learning of time-varying physical parameters: diffusivity of heat diffusion system (a),  propagation speed of wave propagation speed (b).}
  \label{fig:online}
\end{figure*}

\subsection{Burgers' Fluid Flow System}
We consider a two-dimensional system consisting of a moving viscous fluid that follows the Burgers' equation:
\begin{align}
    \frac{\partial u_1}{\partial t} &= -u_1 \frac{\partial u_1}{\partial x} - u_2 \frac{\partial u_1}{\partial y} + \beta \bigg(\frac{\partial^2 u_1}{\partial x^2} \  + \  \frac{\partial^2 u_1}{\partial y^2}\bigg) \ \nonumber \\ &+ \ v_1(x, y, t), \nonumber \\
    \frac{\partial u_2}{\partial t} &= -u_1 \frac{\partial u_2}{\partial x} - u_2 \frac{\partial u_2}{\partial y} + \beta \bigg(\frac{\partial^2 u_2}{\partial x^2} \  + \  \frac{\partial^2 u_2}{\partial y^2}\bigg) \ \nonumber \\ &+ \ v_2(x, y, t), \nonumber \\ 
    & \qquad \qquad (x, y) \in \Omega \subset \mathbb{R}^2, \  t \in [0,T]
    \label{eqn:Burgers}
\end{align}
$\mathbf{u}(x,y,t) = [u_1(x,y,t), u_2(x,y,t)]$ is the flow velocity vector at location $(x,y)$ at time $t$ and, $\mathbf{v}(x, y, t)= [v_1(x,y,t), v_2(x,y,\\t)]$ is the external perturbation vector. $\beta$ is the viscosity of the fluid.
We consider the flow is driven by the pressure gradient caused by an external pressure field. Therefore, the external perturbation $\mathbf{v}$ can be written as $\mathbf{v} = \big[-\frac{\partial P}{\partial x}, -\frac{\partial P}{\partial y}\big]$, where $P(x,y,t)$ denotes the external pressure at location $(x,y)$ at time $t$. We consider the external pressure is varying over time as a high pressure zone and a low pressure zone are moving in circles of random radii in clockwise and counter-clockwise direction, respectively. Figure \ref{fig:pressure_field} shows an example of source maps at the initial and final time step.

Similar to previous two examples, the computation space $\Omega$ is discretized into $64 \times 64$ regular mesh, i.e. $U_t \in \mathbb{R}^{64 \times 64}$. The two pressure zones are Gaussian distributed whose center, peak and spread are chosen randomly for each sequence in the dataset. Training and test dataset are generated using numerical solution method starting from initial condition $U_{t < 0} = 0$ and assuming Neumann boundary condition. Each sequence comprises $200$ frames and the training set contains $300$ such sequences while the test set contains $50$. $20\%$ of the training set is used for validation. Trainable parameters for this system include viscosity $\beta$ and the parameters of residual encoder-decoder network that is used to model the source dynamics. The differential kernels used for this system are $D_{10}, D_{01}, D_{20},$ and $D_{02}$.

Training loss and validation loss of our model for this system are shown in Figure \ref{fig:losses}(c). Figure \ref{fig:be_metrics} shows the quantitative comparison of proposed method with respect to other baselines and choice of hyperparameter $\lambda$ in the task of forecasting and source identification for the Burgers' fluid flow system. Similar to the heat system and wave system, PhICNet shows better accuracy than the PDE-RNN + CNN method for both the forecasting (Figure \ref{fig:be_metrics}(a)) and source identification (Figure \ref{fig:be_metrics}(c)) tasks. Likewise, pure data-driven methods follow the same pattern of the heat system and wave system: RED-Net outperforms ConvLSTM (Figure \ref{fig:be_metrics}(a)). However, unlike the heat system and wave system, the correlation coefficient of the estimated source maps varies significantly for different values of the hyperparameter $\lambda$.
Qualitative comparison of predicted fields by different models along with ground truth is depicted in Figure \ref{fig:be_visual}(a). Figure \ref{fig:be_visual}(b) compares the source pressure gradients predicted by the proposed model and PDE-RNN + CNN with ground truth: PhICNet identifies the source pattern and location correctly while PDE-RNN + CNN fails to capture it after a few initial steps. \\ \\
\textbf{Robustness under noise.} We add Gaussian noise of different variances with the observation to check the robustness of our model in forecasting Burgers' fluid flow system. 
Figure \ref{fig:noise} shows the performance of the PhICNet with noisy observation. Note, we use only few initial observations for prediction. Therefore, SNR is low compared to the no-noise case for the initial few steps, but eventually converges to similar values when the noise level is within the $10\%$ of the variance to dataset. When we add noise having the same variance as the dataset, SNR falls below $0$, as expected.       

\subsection{Online Learning of Time-varying Physical Parameters}
In all aforementioned experiments, we have considered unknown but constant physical parameters which are learned in conjunction with unknown source dynamics. Here, we  investigate if a trained model can adapt with time-varying physical parameter using the method described in section \ref{subsec:online}. %At each time step, the current observation is used to re-tune the physical parameters of the system, while the other parameters are kept frozen.
We vary the diffusivity and propagation speed of the heat system and wave system, respectively, with time and check if the estimated values can track the true profiles. Figure \ref{fig:online} shows the performance of our method in tracking the changes in physical parameter over time. Since wave system is a second order (temporal) system, it is very chaotic compared to the heat system. A small change in propagation speed causes significant error between the true state and the predicted state which helps the model to quickly understand the mismatch and re-learn the propagation speed (Figure \ref{fig:online}(b)). On the other hand, heat diffusion system is relatively slower; when the change in diffusivity is small, the error between the true state and the predicted state is also small. Therefore, the model takes few steps to re-learn the exact diffusivity when the jump in true profile is smaller (Figure \ref{fig:online}(a)).

%% file: conclusion.tex
\section{Conclusion}
\label{sec:conclusion}
We developed a physics-incorporated recurrent neural network PhICNet for spatiotemporal forecasting of dynamical systems with time-varying independent source. Besides forecasting the combined dynamics, our model is also capable of predicting the evolution of source dynamics separately. The current work focuses on presenting the fundamental framework for  spatiotemporal forecasting of partially known dynamical systems with unknown source dynamics in an application-agnostic way. Though we specifically focus on dynamical systems with unobservable source dynamics, our method can potentially be used to predict the evolution of any partially known physical system involving multiple unknown additive dynamics which is a common scenario in many real-world dynamical systems. However, the application of the proposed fundamental method to a specific real-world system will require application-specific modifications. 

An interesting future direction for this work could be to exploit techniques like Neural ODE \cite{chen2018neural}, universal differential equations \cite{rackauckasa2020universal} that not only can improve the memory efficiency, speed and accuracy but also allows learning from irregular observations.

Augmenting physical models in PhICNet improves its generalizability compared to its pure data-driven counterparts and enables long-term prediction on unseen data. However, generalizability of PhICNet is limited by its data-driven component that models the source dynamics. Generalizability of such data-driven models to complex physical dynamics is still not well-understood and requires more in-depth studies.